\UseRawInputEncoding
\pdfoutput=1%For ARXIV

\documentclass[lettersize,journal]{IEEEtran}
\usepackage{amsmath,amsfonts}
\usepackage{algorithmic}
\usepackage{algorithm}
\usepackage{array}
\usepackage[caption=false,font=footnotesize,labelfont=rm,textfont=rm]{subfig}

\usepackage{textcomp}
\usepackage{stfloats}
\usepackage{url}
\usepackage{verbatim}
\usepackage{graphicx}
\usepackage{cite}
\usepackage{color}
\usepackage{doi}
\usepackage{amsmath}

\usepackage{enumitem} % 使用 enumitem 宏包
\usepackage{multirow}

\begin{document}

\title{A Two-stream Hybrid CNN-Transformer Network for Skeleton-based Human Interaction Recognition}

\author{Ruoqi Yin, Jianqin Yin*
        % <-this % stops a space
\thanks{*Corresponding author.}% <-this % stops a space
% \thanks{Manuscript received April 19, 2021; revised August 16, 2021.}
}

\maketitle

\begin{abstract}
Human Interaction Recognition (HIR) is the process of identifying and understanding interactive actions and activities between multiple participants in a specific environment or situation. The aim of this task is to recognise the action interactions between multiple people or entities and their meaning and purpose. Many single Convolutional Neural Network (CNN) has issues, such as the inability to capture global instance interaction features or difficulty in training, leading to ambiguity in action semantics. In addition, the computational complexity of the Transformer cannot be ignored, and its ability to capture local information and motion features in the image is poor. In this work, we propose a Two-stream Hybrid CNN-Transformer Network (THCT-Net), which exploits the local specificity of CNN and models global dependencies through the Transformer. CNN and Transformer simultaneously model the entity, time and space relationships between interactive entities respectively. Specifically, Transformer-based stream integrates 3D convolutions with multi-head self-attention to learn inter-token correlations; We propose a new multi-branch CNN framework for CNN-based streams that automatically learns joint spatio-temporal features from skeleton sequences. The convolutional layer independently learns the local features of each joint neighborhood and aggregates the features of all joints. And the raw skeleton coordinates as well as their temporal difference are integrated with a dual-branch paradigm to fuse the motion features of the skeleton. Besides, a residual structure is added to speed up training convergence. Finally, the recognition results of the two branches are fused using parallel splicing. Multi-grained information modelling is employed to enhance the accuracy and robustness of the action recognition system. Experimental results on diverse and challenging datasets, such as NTU-RGBD, H2O, and Assembly101, demonstrate that the proposed method can better comprehend and infer the meaning and context of various actions, outperforming state-of-the-art methods.
\end{abstract}


\section{Introduction}
Human Interaction Recognition (HIR) has become a significant challenge and research focus in the field of computer vision for identifying and comprehending video content of human actions \cite{ref1,ref2,ref3,ref4}. The rapid development of fields such as social media, intelligent surveillance, and virtual reality has increased the demand for real-time recognition and analysis of human behaviour in videos. The aim of the interactive action recognition task is to extract and recognise human actions from video sequences. These actions may include various activities in daily life, social interactions, or professional actions in specific fields, such as sports or industrial operations \cite{ref5,ref6,ref7,ref8}.

\begin{figure}[htbp]   
  \centering         
  \subfloat[Individual Actions]  
  {
      \label{fig:subfig1}\includegraphics[width=0.15\textwidth]{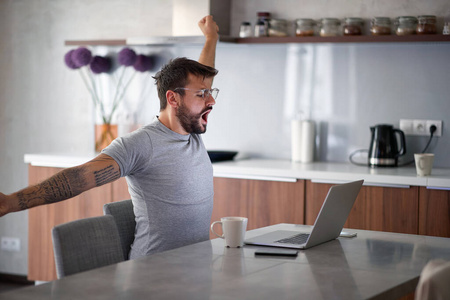}
  }
  \subfloat[Group Activities]
  {
      \label{fig:subfig2}\includegraphics[width=0.15\textwidth]{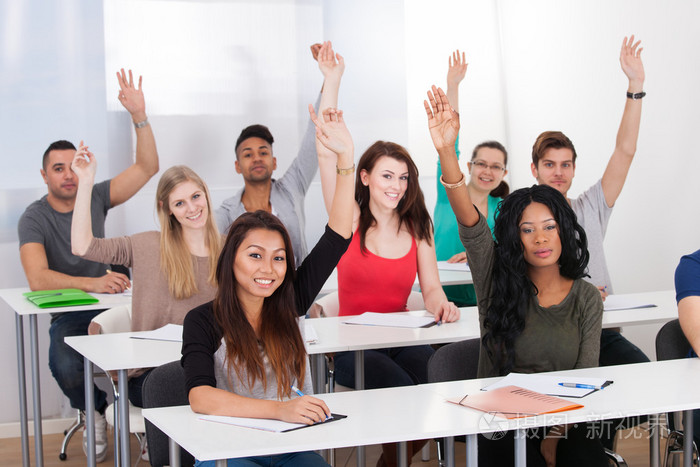}
  }
  \subfloat[Interactive Actions]
  {
      \label{fig:subfig3}\includegraphics[width=0.15\textwidth]{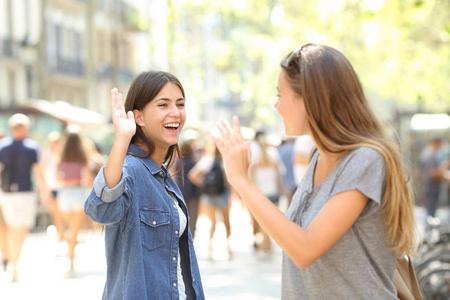}
  }
  \caption{Examples of individual actions (a), group activities (b) and interactive actions (c). (a) Pose of a single person raising his hand could depict the action Stretching. (b) Group activity Answering by raising hands is annotated regardless of the people. (c) In a scene of waving hello, each entity is an integral part of the interactive action.} 
  \label{fig1}          
\end{figure}

The aim of the human interaction recognition task is to identify and comprehend human body movements, gestures, or behaviours, thereby inferring the interaction process between people and objects. Unlike individual actions that relate to the actions of a single subject, and unlike group activities that abstract overall activity events from different individual actions, each goal in an interactive action is essential for explaining the complete semantics. For interactive actions, the term 'individual' refers to the identification of a single action, a sequence of actions, or the behaviour of a specific person. In contrast, the term 'whole' refers to a broader context or scene that contains relationships between multiple individuals or actions. The relationship between these two terms is important because they complement each other and help to more fully understand and explain the meaning and intention of the action. When identifying individual actions, it is important to consider the overall context in order to accurately infer and interpret their meaning. For instance, as shown in Fig. \ref{fig1}, while a single person raising their hand in a video may seem like a simple action, understanding that it occurs during a greeting scene provides a much richer understanding of its significance. The relationships and interactions among multiple individuals are integral to the overall situation. Identifying these interactive actions aids in comprehending the purpose and significance of individual actions within the broader scenario. For instance, in a social setting, a sequence of actions may comprise a conversational exchange, and recognising the overall interaction can reveal meaning and emotion beyond words.

In this task, videos are considered as spatio-temporal sequences that contain rich information. Each frame represents a moment, and the sequence represents the evolution of these moments on the timeline. Therefore, accurately understanding and identifying actions in videos requires not only modeling spatial information, such as posture and object location, but also capturing and understanding temporal information, which is the evolution of actions.

The challenges of this complex task are twofold. Firstly, video sequences are typically high-dimensional and contain a vast amount of information. Secondly, there are various types of interactive actions, including people-to-people, hands-to-hands, and hands-to-objects. Different interacting entities have distinct physical structures and interaction modes, resulting in complexity and variability in interaction modelling. To tackle these challenges, researchers have focused on developing different computational models and techniques to achieve human interaction recognition \cite{ref9,ref10,ref11,ref12}. Among these methods, Convolutional Neural Networks (CNNs) are commonly used to extract spatial information from video frames and capture static features frame by frame. However, modelling the temporal dependence of long sequences has always been a challenge. The emergence of deep learning technology has led to the development of Transformer, a sequence modeling tool that utilizes self-attention mechanism to achieve better modeling of long-term dependencies. As a result, Transformer has been introduced into human interaction recognition tasks, demonstrating great potential in capturing long sequence temporal relationships and modeling action sequences.

The Vision-Transformer (ViT) is a vision model that is entirely based on the Transformer structure. In comparison to traditional CNN vision models, ViT has shortcomings in both model structure and feature representation. Specifically, ViT divides the image into several fixed-size patches and subsequently performs feature extraction and classification on them. However, this will cause the original ViT model to be sensitive to the size of the input image, which can limit its ability to utilize global image information and ultimately affect its performance. It is important to maintain a balanced approach to classification performance. Secondly, the original ViT does not include multi-layer convolution and pooling operations in CNN. This may limit its ability to extract certain image features, resulting in a weaker ability to extract detailed information such as texture and shape. CNN performs well in image processing and can handle complex image features, particularly local features. However, its performance is weaker when processing global information. In contrast, Transformer excels in the NLP field, particularly in modeling and generating sequence data, and has an advantage in processing global information. By combining CNN with Transformer, this model can effectively capture and process both local and global information in images, resulting in improved performance.

Overall, there have been some excellent works on interactive action recognition methods based on CNN or Transformer. However, there is still room for improvement. To address these issues and combine the advantages of CNN and Transformer networks, we propose a Two-stream Hybrid CNN-Transformer Network (THCT-Net), which exploits the local specificity of CNN and models global dependencies through the Transformer. CNN and Transformer simultaneously model the entity, time and space relationships between interactive entities respectively. Specifically, Transformer-based stream integrates 3D convolutions with multi-head self-attention to learn inter-token correlations; We propose a new multi-branch CNN framework for CNN-based streams that automatically learns joint spatio-temporal features from skeleton sequences. The convolutional layer independently learns the local features of each joint neighborhood and aggregates the features of all joints. And the raw skeleton coordinates as well as their temporal difference are integrated with a dual-branch paradigm to fuse the motion features of the skeleton. Besides, a residual structure is added to speed up training convergence. Finally, the recognition results of the two branches are fused using parallel splicing. Experiments on three popular datasets verify that this model has the best fusion effect.

The main contributions of this paper are as follows：

1) We propose a new Two-stream Hybrid CNN-Transformer Network (THCT-Net) for human interaction recognition tasks, which uses Transformer self-attention module and traditional convolutional layers to learn multi-granularity context.

2) We propose a new multi-branch CNN framework that automatically learns joint spatio-temporal features from skeleton sequences. The convolutional layer independently learns the local features of each joint neighborhood and aggregates the features of all joints. And the raw skeleton coordinates as well as their temporal difference are integrated with a dual-branch paradigm to fuse the motion features of the skeleton. Besides, a residual structure is added to speed up training convergence.

3) Extensive experiments on NTU RGB+D 120, H2O and Assembly101 datasets consistently verify the effectiveness of our method, which outperforms most interactive action recognition methods.

\section{Related Work}
{\bf{Human Interaction Recognition}}. For tasks involving human interaction recognition, TA-GCN \cite{ref9} uses topology-aware graph convolutional networks to learn the interdependencies and connections between different graph entities. It also computes the topology of multi-graph structures to learn the interdependence between the two hands and objects. LSTM-IRN \cite{ref10} exploits minimal prior knowledge about human body structure, uses different body parts in posture information as independent objects, and performs pairwise modeling of their relationships. Raptis et al. \cite{ref11} propose to cast the learning in a max-margin discriminative framework where treat keyframes as latent variables. This allows model to jointly learn a set of the most discriminative keyframes while also learning the local temporal context between them. 

IGFormer \cite{ref12} is the first network to adopt a Transformer-based architecture and utilise prior knowledge of human body structure to design interactions. It builds interaction graphs based on semantic and distance correlations between interacting body parts and enhances each person's representation by aggregating information of interacting body parts based on the learning graph. ISTA-Net \cite{ref13} does not require subject-type-specific graph prior knowledge to model diverse interacting entities. By extending an additional entity dimension in attention tokens, it can simultaneously and also effectively capture interactive and spatiotemporal correlations of interactive actions.

In summary, there have been some excellent works for human interaction recognition tasks, each demonstrating their respective advantages. For instance, CNN extracts features through shared convolution kernels, which reduces the number of network parameters, improves model efficiency, and provides translation invariance. However, it has a limited receptive field. Subsequently, the Long Short-term Memory Network (LSTM) \cite{ref14} gained popularity as a model for individual dynamics in single-person action recognition due to its capacity to capture temporal motion information within a specific range. However, existing Recurrent Neural Networks (RNNs) only concentrate on capturing the dynamics of human interactions by merely combining all individual dynamics or modelling them as a whole, disregarding the interconnected dynamics of how human interactions evolve over time. Vision Transformer (ViT) \cite{ref15} uses a pure Transformer structure to replace CNN, enabling it to capture global information of an image and surpassing the CNN structure in many visual tasks. This paper aims to explore a new human interaction recognition model that effectively combines the advantages of previous work and further improves recognition performance.

{\bf{Hybrid CNN-Transformer}}. In recent years, research on hybrid CNN-Transformer models in computer vision has become a hot topic. This model combines the advantages of both CNN and Transformer to improve performance in various computer vision tasks. The success of CNN is due to its inherent inductive biases, namely translation invariance and local correlation. However, the limited receptive field of CNN makes it difficult to capture global information. In contrast, Transformer can capture long-distance dependencies. Therefore, after the emergence of ViT, many works have attempted to combine CNN and Transformer. This allows the network structure to inherit the advantages of both CNN and Transformer, retaining global and local features to the greatest extent possible.

Theoretically, Transformers can achieve better model performance than CNNs. However, calculating global attention results in significant computational losses, particularly in shallow networks. The computational complexity increases with the size of the feature map. Therefore, some methods propose inserting the Transformer into the CNN backbone network or using a Transformer module to replace a specific convolution module.  BoTNet \cite{ref16} utilises Multi-Head Self-Attention to replace the $3\times3$ convolution in ResNet Bottleneck, resulting in a new network structure called Bottleneck Transformer. This approach combines the local features of CNN with the overall image focusing features of Transformer, while also significantly reducing computational requirements.

CNN exhibits locality and translation invariance. Locality pertains to adjacent points in the feature map, while translation invariance involves using the same matching rules for different regions. While the inductive bias of CNN enhances its performance on small data sets, it can limit its performance on larger ones. Consequently, some researchers have attempted to incorporate the inductive bias of CNN into Transformers to expedite network convergence. To decrease ViT's reliance on vast amounts of data, Touvron et al. \cite{ref17} proposed the Data-efficient Image Transformer (DeIT). This approach enhances the network's performance on small data sets by utilizing data augmentation and regularization techniques. Additionally, a distillation strategy is introduced, which employs a teacher network to guide the student network. Dai et al. \cite{ref18} proposed CoAtNet, a Convolution and Attention Network that incorporates depth convolution into the attention module. In depth convolution, each convolution kernel is responsible for one channel, resulting in lower parameters and operation costs compared to normal convolution. In depth convolution, each convolution kernel is responsible for one channel, resulting in lower parameters and operation costs compared to normal convolution. CoAtNet employs shallow networks with stacked convolutional layers. However, we discovered that hybrid models for human interaction recognition are extremely rare. Therefore, in this work, we introduce a two-stream hybrid CNN-Transformer network to enhance the generalisation ability and convergence speed of the model through parallel splicing. 

\section{METHODOLOGY}
\begin{figure*}[!t]
\centering
\includegraphics[width=7.1in]{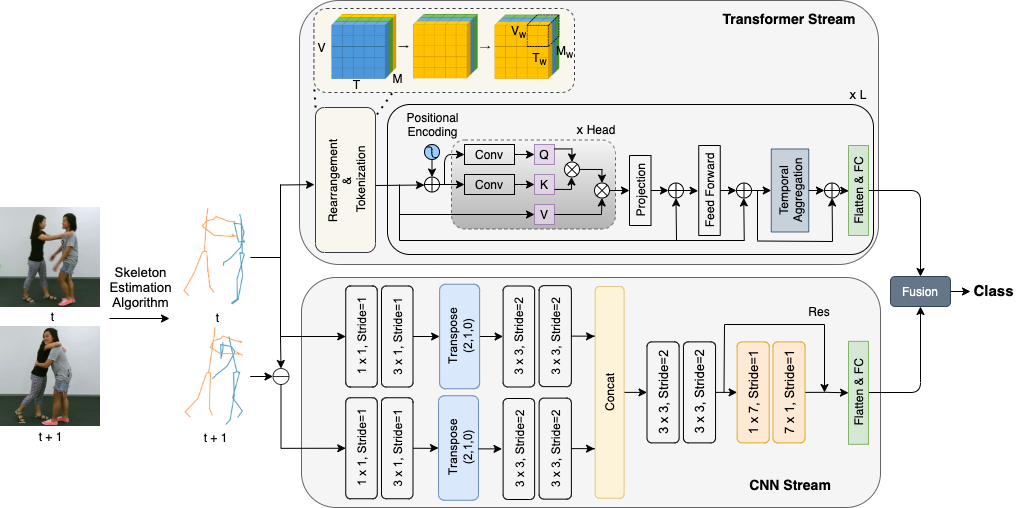}%
\caption{The overall architecture of the proposed THCT-Net for skeleton-based human interaction recognition.}
\label{fig2}
\end{figure*}

As shown in Fig. \ref{fig2}, THCT-Net consists of two parallel streams processing information differently: 1) CNN stream, which learns joint spatiotemporal features from skeleton sequences. The convolutional layer independently learns the local features of each joint neighborhood and aggregates the features of all joints. And the raw skeleton coordinates as well as their temporal difference are integrated with a dual-branch paradigm to fuse the motion features of the skeleton. Besides, a residual structure is added to speed up training convergence. 2) Transformer branch, where it integrates 3D convolutions with multi-head self-attention to learn inter-token correlations. The benefit of the proposed branch-in-parallel approach: by leveraging the merits of CNNs and Transformers, we argue that THCT-Net can capture global information while preserving sensitivity on low-level context.

\subsection{Transformer Stream}
The design of Transformer stream follows ISTA-Net \cite{ref13}. The input skeleton sequence ${X}_{input} \in∈ {\mathbb{R}}^{3\times T\times V\times M}$ is defined based on the estimated 3D skeleton of $M$ interactive entities interacting within time $T$, with each entity containing $V$ joints.

The first step is to rearrange the input entities. When dealing with interactive entities, some are semantically unordered and interchangeable, such as people. Therefore, they can be arranged in any order while still representing the same interaction. The input skeleton sequence of size $C\times T\times V\times M$ is divided into $M$ parts along the interaction dimension, with each part representing the joint motion of a body. This is achieved through the following equation: 
\begin{equation}
\begin{aligned}
    [{X}_1, {X}_2,..., {X}_i ,..., {X}_{M}] = Split({X}_{input}),
\end{aligned}
\end{equation}
where [1, 2, ..., $i$, ..., $M$] represents the index of position order along the interaction dimension. 

We could rearrange the original ${X}_{input}$ as follows:
\begin{equation}
\begin{aligned}
    {\widetilde{X}}_{input} = Concat([{X}_{{a}_{1}}, {X}_{{a}_{2}},..., {X}_{{a}_{i}} ,..., {X}_{{a}_{M}}]),
\end{aligned}
\end{equation}
where $[{a}_{1}, {a}_{2}, ..., {a}_{i}, ..., {a}_{M}]$ is an arbitrary arrangement of indexes [1, 2, ..., $i$, ..., $M$].

The input permutation ${\widetilde{X}}_{input}$ is selected in each training epoch, while the original input ${X}_{input}$ is used in validation and testing.

Subsequently, the skeleton tensors are tokenized via a 3D sliding window to obtain interactive spatiotemporal tokens. Non-overlapping 3D windows are used to partition the input data. A window $W$ of size ${T}_{w}\times {J}_{w}\times {E}_{w}$ slides along the temporal, spatial, and interaction dimensions. The input of size $C\times T\times V\times M$ is divided into $U = [T/{T}_{w}]\times [V/{V}_{w}]\times [M/{M}_{w}]$ patches of size $C\times {T}_{w}\times {V}_{w}\times {M}_{w}$ in total. They represent interactive spatiotemporal local features for interactive skeleton sequences. The 3D convolution operation, followed by the batch normalization and an activation function, serves as the embedding layer for interactive spatiotemporal tokens.

Then these tokens are fed to $L$ Multi-head Self-attention Blocks to learn high-level cross frame, joint and subject representations. Similar to standard multi-head self-attention, the input ${X}_{L_{i-1}}$ undergoes transformation into multiple sets of queries $Q$, keys $K$, and values $V$ as follows:
\begin{equation}
\begin{aligned}
    Q = Conv3D_{(1\times 1\times 1)}({X}_{L_{i-1}}+PE({X}_{L_{i-1}})),
\end{aligned}
\end{equation}
\begin{equation}
\begin{aligned}
    K = Conv3D_{(1\times 1\times 1)}({X}_{L_{i-1}}+PE({X}_{L_{i-1}})),
\end{aligned}
\end{equation}
\begin{equation}
\begin{aligned}
    V = {X}_{L_{i-1}},
\end{aligned}
\end{equation}
where positional encoding implemented with circular functions is $PE(\cdot)$. The number of sets, namely heads, is denoted as $H$.

Self-attention scores ${X}_{{L}_{i}}^{h}$ of the $h$-th head could be calculated as:
\begin{equation}
\begin{aligned}
    {X}_{{L}_{i}}^{h} = (\alpha tanh(\frac{QK
    ^{T}}{\sqrt{C_{\beta}}})+A)V,
\end{aligned}
\end{equation}
where ${QK^{T}}$ is divided by the square root of the feature length ${C_{\beta} = T_{w}\times V_{w}\times M_{w}\times C_{{L}_{i}-qkv}}$. A trainable regularized matrix ${A} \in∈ {\mathbb{R}}^{U\times U}$ is added to the normalized attention map with a trainable balanced factor $\alpha$, which can benefit correlation learning \cite{ref22,ref23}. All scores ${X}_{{L}_{i}}^{h}$ of $H$ heads are concatenated to get ${X}_{{L}_{i}}^{H}$. 

A 3D $1\times 1\times 1$ convolution with residual connections implements the feed forward network (FFN). The last component is the temporal aggregation layer, it uses 3D convolution with kernel size 5 in the temporal dimension to aggregate sequence features. Prediction is finally made through Global Average Pooling (GAP) following with a fully connected (FC) layer.

\subsection{CNN Stream}
Convolutional Neural Networks (CNNs) have been highly successful in the field of deep learning and have played a crucial role in tasks such as image recognition and computer vision. Unlike sequential structures such as RNNs, CNNs can encode spatial and temporal context information simultaneously.

This section provides a detailed description of the proposed CNN framework, which aims to learn both the spatial global features and temporal evolution of skeleton sequences. Fig. \ref{fig2} displays the network architecture of the proposed framework. The skeleton sequence $X$ can be represented by a $C\times T\times V\times M$ tensor, where $C$ represents the coordinate dimension of the joints (e.g. 3 for a 3D skeleton: $x,y,z$), $T$ represents the number of frames in the sequence, $V$ represents the number of joints in the skeleton, and $M$ represents the number of people. For interactive actions, activities such as hugging and shaking hands require the participation of multiple people. To ensure scalability in multi-person scenarios, we utilize early fusion to aggregate the joint points of all individuals. This involves stacking all joints from multiple individuals as the input of the network, resulting in a tensor of size $(C, T, V\times M)$ to represent the input skeleton sequence.

Firstly, we encode the data using convolutional layers with kernel sizes of $1\times1$ and $3\times1$. By keeping the kernel size along the joint dimension to 1, the model is forced to learn point-level representations independently from the 3D coordinates of each joint. It is observed that the output of the convolutional layer represents the global response of all input channels. If a 3D tensor $F$ is represented as ${d}_{1}\times {d}_{2}\times {d}_{3}$, with dimension ${d}_{i}$ specified as a channel and the other two dimensions encoding local context, any information from dimension ${d}_{i}$ can be globally aggregated. This allows for the assignment of different contexts by transposing the tensor. Previous CNN-based methods have specified joint coordinates as channels to learn local features of each joint neighbourhood \cite{ref19,ref20,ref21} , which may result in the inability to capture some long-range joint interaction information. Therefore, the feature map is transposed using the parameters (2,1,0) to move the joint dimensions to the channels of the tensor, i.e. $(V\times M, T, C)$. If we treat each joint of the skeleton as a channel, the convolutional layer can learn the global features of all joints more easily.

In addition to learning the spatial global features of skeleton sequences, it is important to consider the inter-frame representation of the skeleton's temporal evolution as a clue for identifying potential actions. Motion information is introduced by calculating the difference between frames, which enhances action information in time series data. This approach helps to better capture and represent the characteristics of actions. An additional branch is introduced to learn skeleton motion information from an $C\times T\times V\times M$ tensor. For the skeleton of a person in frame $t$, we formulate it as ${S}^{t} =\{{J}_{1}^{t}, {J}_{2}^{t}, ..., {J}_{V}^{t}\}$ where $V$ is the number of joint and $J = (x,y,z)$ id a 3D joint coordinate. The skeleton motion is defined as the temporal difference of each joint between two consecutive frames:
\begin{equation}
\begin{aligned}
	{M}^{t} & = {S}^{t+1} - {S}^{t}\\
            & = \{{J}_{1}^{t+1} - {J}_{1}^{t}, {J}_{2}^{t+1} - {J}_{2}^{t}, ..., {J}_{V}^{t+1} - {J}_{V}^{t}\}.
\end{aligned}
\end{equation}

The network processes the raw skeleton coordinates $S$ and the skeleton motion $M$ independently using a dual-branch paradigm. Both branches share the same architecture. However, their parameters are learned separately. After passing through the $(64, 3\times3)$ convolutional layer, the feature maps of the two branches are fused by concatenating across channel dimensions. The fused features learn a richer feature representation through a residual module. By combining $1\times7$ and $7\times1$ size convolutions, convolution operations can be performed on the input tensor in different directions, which is equivalent to using a larger size convolution kernel. This approach captures spatial and cross-channel information more efficiently. Residual connections enable the direct transfer of the difference between input and output, expanding the network's receptive field and allowing it to capture a wider range of spatial information and semantic features. Additionally, reducing the model's parameters helps mitigate the risk of overfitting and improves its trainability. Finally, the feature maps are flattened into vectors and passed through two fully connected layers for the final classification.

Finally, we late-fuse the classification scores of the two streams in a weighted manner to obtain the final recognition result.

\section{EXPERIMENTS}

\begin{table*}[]
\centering
\caption{COMPARISONS OF ACTION RECOGNITION METHODS ON THREE DIFFERENT INTERACTIVE ACTION DATASETS}
\setlength{\tabcolsep}{14pt}
\renewcommand{\arraystretch}{1.35}
\label{tab1}
\begin{tabular}{c|c|c|cc|c|c}
\hline
\multirow{2}{*}{Type}        & \multirow{2}{*}{Methods} & \multirow{2}{*}{Year} & \multicolumn{2}{c|}{\begin{tabular}[c]{@{}c@{}}NTU RGB+D 120 \\ - 26 Mutual Actions(\%)\end{tabular}} & \multirow{2}{*}{H2O(\%)} & \multirow{2}{*}{Assembly101(\%)} \\ \cline{4-5}
                             &                          &                       & \multicolumn{1}{c|}{X-Sub}                                   & X-Set                                  &                          &                                  \\ \hline
\multirow{7}{*}{LSTM}        & Co-LSTM \cite{ref27}                  & AAAI 2016             & \multicolumn{1}{c|}{-}                                       & -                                      & -                        & -                                \\
                             & ST-LSTM \cite{ref28}                  & ECCV 2016             & \multicolumn{1}{c|}{63.00}                                   & 66.60                                  & -                        & -                                \\
                             & GCA \cite{ref29}                      & CVPR 2017             & \multicolumn{1}{c|}{70.60}                                   & 73.70                                  & -                        & -                                \\
                             & VA-LSTM \cite{ref30}                  & ICCV 2017             & \multicolumn{1}{c|}{-}                                       & -                                      & -                        & -                                \\
                             & 2s-GCA \cite{ref31}                   & TIP 2018              & \multicolumn{1}{c|}{73.00}                                   & 73.30                                  & -                        & -                                \\
                             & H+O \cite{ref32}                      & CVPR 2019             & \multicolumn{1}{c|}{-}                                       & -                                      & 68.88                    & -                                \\
                             & LSTM-IRN \cite{ref10}                 & TMM 2022              & \multicolumn{1}{c|}{77.70}                                   & 79.60                                  & -                        & -                                \\ \hline
\multirow{10}{*}{GCN}        & ST-GCN \cite{ref33}                   & AAAI 2018             & \multicolumn{1}{c|}{78.90}                                   & 76.10                                  & 73.76                    & -                                \\
                             & AS-GCN \cite{ref34}                   & CVPR 2019             & \multicolumn{1}{c|}{82.90}                                   & 83.70                                  & -                        & -                                \\
                             & 2s-AGCN \cite{ref35}                  & CVPR 2019             & \multicolumn{1}{c|}{-}                                       & -                                      & -                        & 26.70                            \\
                             & MS-G3D \cite{ref36}                   & CVPR 2020             & \multicolumn{1}{c|}{-}                                       & -                                      & -                        & 26.86                            \\
                             & CTR-GCN \cite{ref37}                  & ICCV 2021             & \multicolumn{1}{c|}{89.32}                                   & 90.19                                  & -                        & 26.25                            \\
                             & TA-GCN \cite{ref9}                   & ICCV 2021             & \multicolumn{1}{c|}{-}                                       & -                                      & 79.25                    & -                                \\
                             & LST \cite{ref38}                      & arXiv 2022            & \multicolumn{1}{c|}{89.27}                                   & 90.60                                  & -                        & -                                \\
                             & TCA-GCN \cite{ref39}                  & arXiv 2022            & \multicolumn{1}{c|}{88.37}                                   & 89.30                                  & -                        & -                                \\
                             & HD-GCN \cite{ref40}                   & arXiv 2022            & \multicolumn{1}{c|}{88.25}                                   & 90.08                                  & -                        & -                                \\
                             & InfoGCN \cite{ref41}                  & CVPR 2022             & \multicolumn{1}{c|}{90.22}                                   & 91.13                                  & -                        & 25.63                            \\ \hline
\multirow{4}{*}{Transformer} & DSTA-Net \cite{ref22}                 & ACCV 2020             & \multicolumn{1}{c|}{88.92}                                   & 90.10                                  & -                        & -                                \\
                             & STSA-Net \cite{ref23}                 & Neurocomputing 2023   & \multicolumn{1}{c|}{90.20}                                   & 90.97                                  & -                        & -                                \\
                             & IGFormer \cite{ref12}                 & ECCV 2022             & \multicolumn{1}{c|}{85.40}                                   & 86.50                                  & -                        & 22.33                            \\
                             & ISTA-Net \cite{ref13}                 & IROS 2023             & \multicolumn{1}{c|}{90.56}                                   & 91.72                                  & 89.09                    & 28.01                            \\ \hline
CNN-Transformer              & THCT-Net (Ours)          & 2023                  & \multicolumn{1}{c|}{\textbf{91.00}}                                   & \textbf{91.86}                                  & \textbf{92.98}                    & \textbf{28.42}                            \\ \hline
\end{tabular}
\end{table*}

\subsection{Datasets}
The detailed descriptions of three public datasets are as follows:

\begin{itemize}[label=$\bullet$, leftmargin=*] % 设置项目符号为圆点，leftmargin=* 去除默认的缩进
  \item \textbf{NTU RGB+D 120} \cite{ref24},  the extension version of \textbf{NTU RGB+D} \cite{ref25}, is a widely-used action recognition dataset. It provides 114,480 samples of 120 human actions. In our experiments we focus on a subset of \textbf{NTU RGB+D 120} Dataset, which consists of 26 kinds of mutual actions (named \textbf{NTU Mutual}, for short).
  \item \textbf{H2O} \cite{ref9} is the first dataset constructured for egocentric 3D interaction recognition. The images of the H2O dataset are acquired in indoor settings in which the subjects interact with eight different objects using both of their hands. The dataset includes 571,645 RGBD frames, and features four participants performing 36 distinct action classes in three different environments. With 3D pose of both hands and pose of manipulated objects, H2O dataset facilitates hand-to-hand and hand-to-object interactions understanding.
  \item \textbf{Assembly101} \cite{ref26}  is a large procedural activity dataset. 3D hand poses are provided to advance 3D interaction recognition from egocentric views. It’s a tough task due to the dataset’s complexity, which includes over 1,300 fine-grained classes of hand-to-object interactions. Each class consists of a single verb and an object that is manipulated. Additionally, the absence of object poses adds another layer of difficulty to judging the interactive actions.

\end{itemize}

Statistics and difficulties of these datasets are summarized in Table I and Fig. 3. For evaluation on NTU Mutual, we employ the Cross-subject (X-Sub) and Cross-set (X-Set) criteria \cite{ref24}, using only the joint modality to ensure fair comparisons without fusion. For H2O and Assembly101, we follow the training, validation, and test splits described in \cite{ref9} and \cite{ref26}, respectively.

\subsection{Implementation Details}
All of our experiments are conducted on a machine equipped with four NVIDIA GeForce RTX 3090 GPUs and CUDA version 12.2. For training on NTU Mutual dataset, SGD optimizer is used with Nesterov momentum of 0.9, a initial learning rate of 0.1 and a decay rate 0.1. Window size is set to $[20, 1, 2]$. Cross entropy is used as loss function with label smoothing factor 0.1 and temperature factor 1.0. Batch size is 32. Each training process was terminated after 110 epochs.

\subsection{Results and Analyses}
\subsubsection{Comparison with Baselines}
We used the Transformer-based method ISTA-Net \cite{ref13} as the baseline. Table \ref{tab1} shows the recognition accuracy of the proposed THCT-Net is better than baseline on three datasets. CNN is effective at extracting local image features, with superior generalization ability and faster convergence speed. On the other hand, Transformer excels at capturing global semantic information and can produce excellent results on large datasets. Concatenating CNN and Transformer models in parallel can lead to better performance compared to using either model alone.

\subsubsection{Comparison with Related Methods}
Table \ref{tab1} reports the experimental results on NTU Mutual, H2O and Assembly101 datasets. The proposed THCT-Net outperforms many LSTM-, GCN-, Transformer-based action recognition methods and other human interaction recognition methods. THCT-Net achieves 0.44\%, 0.14\%, 4.07\% and 0.41\% gains over the most related interactive action recognition method, ISTA-Net \cite{ref13}, on NTU Mutual X-Sub, X-Set, H2O and Assembly101. THCT-Net also outperforms InfoGCN \cite{ref41} by 0.78\% and 0.73\% on NTU Mutual, TA-GCN \cite{ref9} by 13.73\% on H2O, and MS-G3D \cite{ref36} by 1.56\% on Assembly101. THCT-Net utilises the local specificity of CNNs and models global dependencies through the use of a transformer. The CNNs and transformer work together to model the physical, temporal, and spatial relationships between interacting entities. The recognition results from both branches are then combined through concurrent splicing to improve accuracy and robustness by modelling information at multiple granularities.

\section{Conclusion}
For the human interaction recognition task, we propose a Two-stream Hybrid CNN-Transformer network (THCT-Net). The CNN models the temporal relationships between entities, while the Transformer models the spatial relationships between interacting entities. This approach mitigates the problem of ambiguity in the semantics of actions caused by a single model. Specifically, Transformer-based stream integrates 3D convolutions with multi-head self-attention to learn inter-token correlations; We propose a new multi-branch CNN framework for CNN-based stream that automatically learns joint spatio-temporal features from skeleton sequences. The convolutional layer independently learns the local features of each joint neighborhood and aggregates the features of all joints. And the raw skeleton coordinates as well as their temporal difference are integrated with a dual-branch paradigm to fuse the motion features of the skeleton. Besides, a residual structure is added to speed up training convergence. Finally, the recognition results of the two branches are fused using parallel splicing. Multi-grained information modelling is employed to enhance the accuracy and robustness of the action recognition system. Extensive experiments on NTU RGB+D 120, H2O and Assembly101 datasets consistently verify the effectiveness of our method, which outperforms most interactive action recognition methods.

\section*{Acknowledgments}
This work was supported partly by the National Natural Science Foundation of China (Grant No. 62173045, 62273054), partly by the Fundamental Research Funds for the Central Universities (Grant No. 2020XD-A04-3), and the Natural Science Foundation of Hainan Province (Grant No. 622RC675).

%\newpage

\vspace{11pt}

\vspace{-33pt}
\begin{IEEEbiography}[{\includegraphics[width=1in,height=1.2in,clip,keepaspectratio]{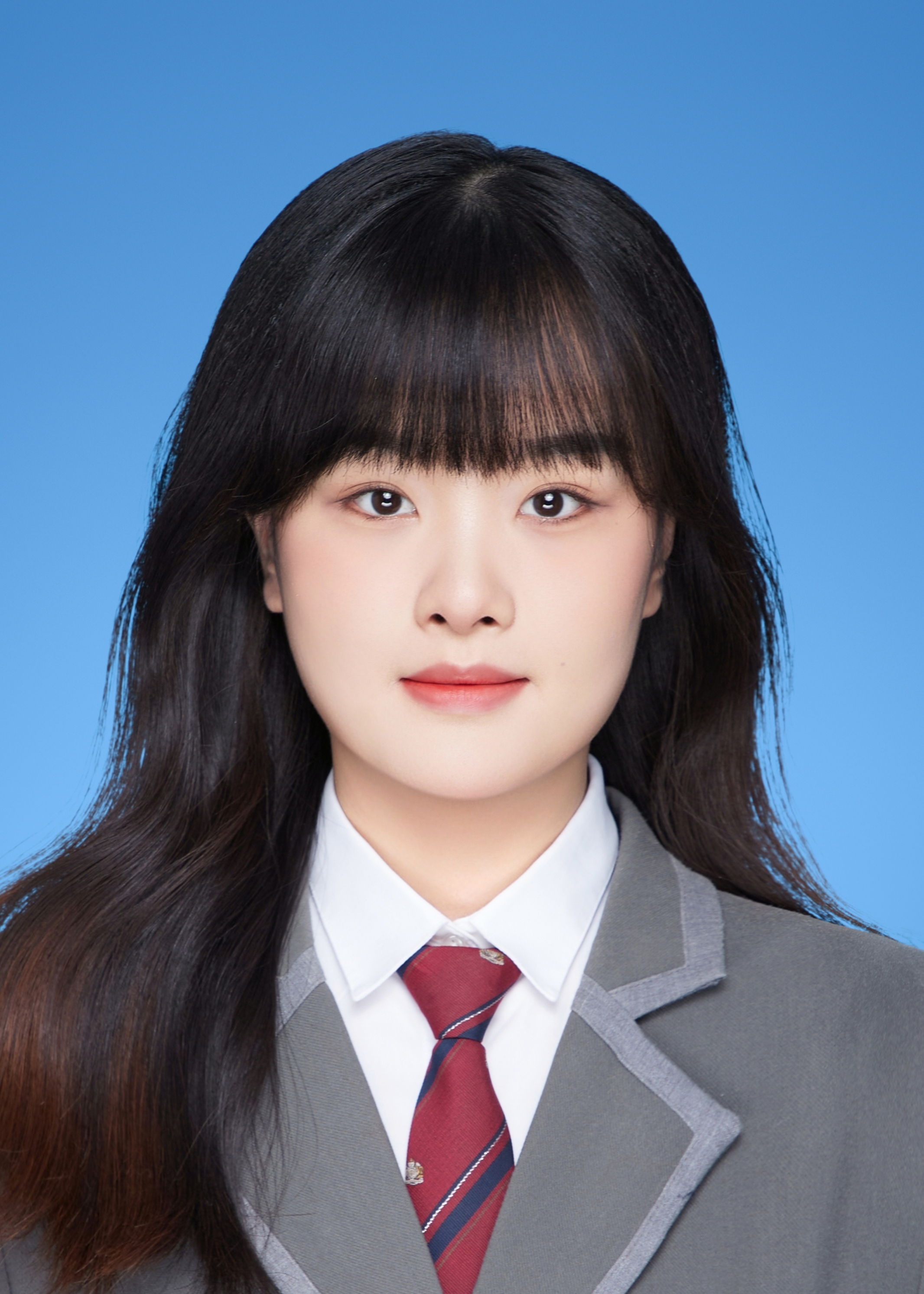}}]{Ruoqi Yin}
She currently is a bachelor in Artificial
Intelligence School, Beijing University of Posts and
Telecommunications, Beijing, China. Her research
interests include image processing, pose estimation, and deep learning.
\end{IEEEbiography}

\begin{IEEEbiography}
[{\includegraphics[width=1in,height=1.2in,clip,keepaspectratio]{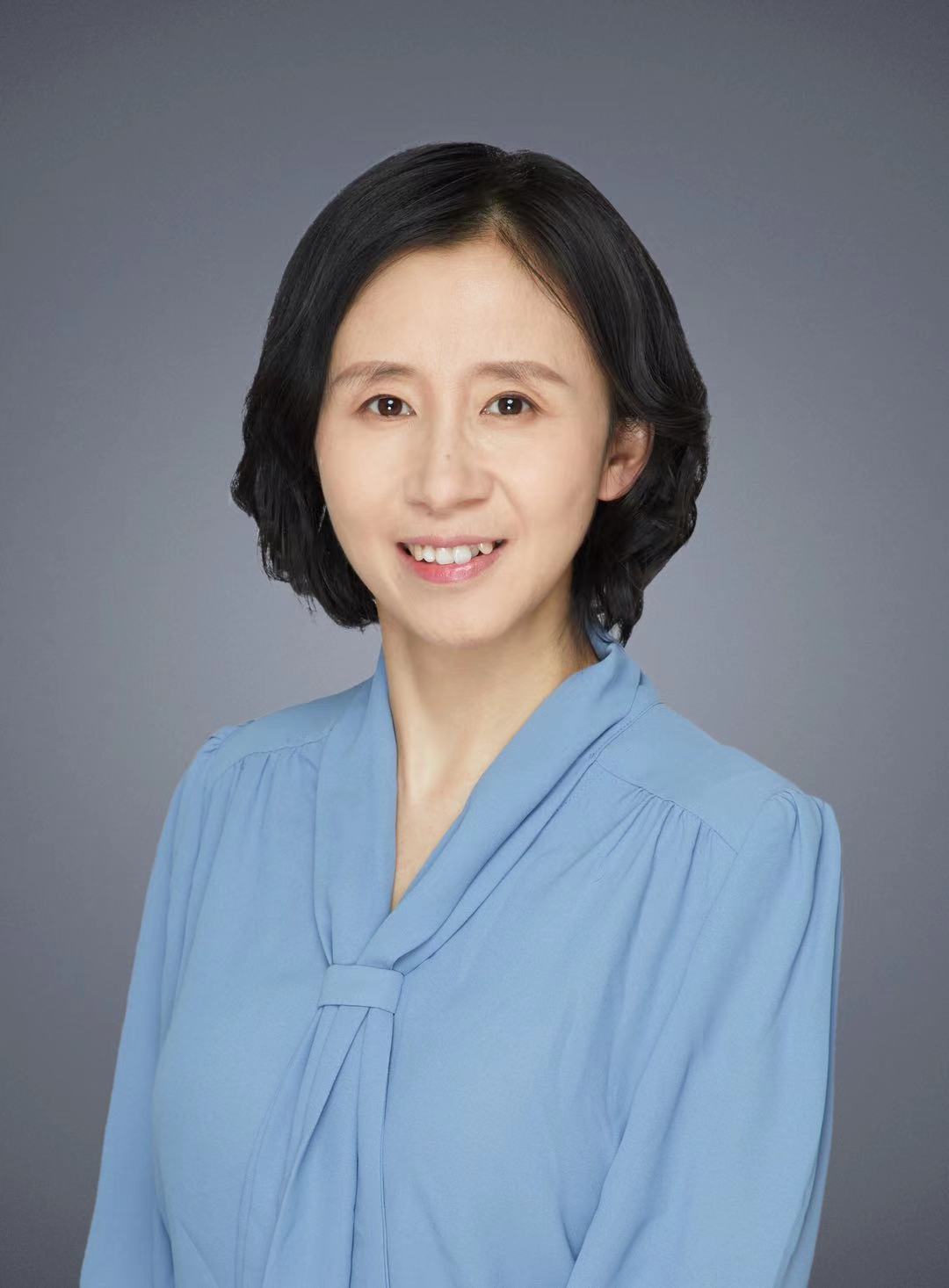}}]{Jianqin Yin}
(Member, IEEE) received the Ph.D. degree from Shandong University, Jinan, China, in 2013. She is currently a Professor with the School of Artificial Intelligence, Beijing University of Posts and Telecommunications, Beijing, China. Her research interests include service robot, pattern recognition, machine learning, and image
processing.
\end{IEEEbiography}

\vspace{6pt}

\vfill
\end{document}